\documentclass[conference]{IEEEtran}
\IEEEoverridecommandlockouts
\usepackage{cite}
\usepackage{amsmath,amssymb,amsfonts}
\usepackage{algorithmic}
\usepackage{graphicx}
\usepackage{textcomp}
\usepackage{xcolor}
\usepackage{graphicx}
\usepackage{colortbl}
\definecolor{silver}{rgb}{0.753,0.753,0.753}
\definecolor{mgray}{rgb}{0.128,0.128,0.128}

\def\BibTeX{{\rm B\kern-.05em{\sc i\kern-.025em b}\kern-.08em
    T\kern-.1667em\lower.7ex\hbox{E}\kern-.125emX}}

\usepackage{fancyhdr}
\begin{document}

\makeatletter
\newcommand{\linebreakand}{%
  \end{@IEEEauthorhalign}
  \hfill\mbox{}\par
  \mbox{}\hfill\begin{@IEEEauthorhalign}
}
\makeatother

\title{Fine-grained Affective Processing Capabilities Emerging from Large Language Models\\
%\thanks{This research is partly sponsored by the Hybrid Intelligence project.}
}

 \author{\IEEEauthorblockN{1\textsuperscript{st} Joost Broekens}
 \IEEEauthorblockA{\textit{LIACS} \\
 \textit{Leiden University}\\
 Leiden, The Netherlands \\
 joost.broekens@gmail.com}
 \and
 \IEEEauthorblockN{2\textsuperscript{nd} Bernhard Hilpert}
 \IEEEauthorblockA{\textit{LIACS} \\
 \textit{Leiden University}\\
 Leiden, The Netherlands \\
 b.hilpert@liacs.leidenuniv.nl}
 \and
 \IEEEauthorblockN{3\textsuperscript{rd} Suzan Verberne}
 \IEEEauthorblockA{\textit{LIACS} \\
 \textit{Leiden University}\\
 Leiden, The Netherlands \\
 s.verberne@liacs.leidenuniv.nl}
 \and
 \IEEEauthorblockN{4\textsuperscript{th} Kim Baraka}
 \IEEEauthorblockA{\textit{Department of Computer Science} \\
 \textit{Free University Amsterdam}\\
 Amsterdam, the Netherlands \\
 k.baraka@vu.nl}
 \linebreakand
 \IEEEauthorblockN{5\textsuperscript{th} Patrick Gebhard}
  \IEEEauthorblockA{\textit{German Research Center}\\ \textit{for Artificial Intelligence (DFKI)} \\
 Saarbruecken, Germany \\
 patrick.gebhard@dfki.de}
 \and
 \IEEEauthorblockN{6\textsuperscript{th} Aske Plaat}
 \IEEEauthorblockA{\textit{LIACS} \\
 \textit{Leiden University}\\
 Leiden, The Netherlands \\
 aske.plaat@gmail.com}
 }

\maketitle
\thispagestyle{fancy}
%Submissions may be up to 8 pages (7 pages + 1 page for Ethical Impact Statement and references) in the conference paper format. % https://acii-conf.net/2023/wp-content/uploads/2023/03/2023-ACII-Submission-Guidelines.pdf

\begin{abstract}
Large language models, in particular generative pre-trained transformers (GPTs), show impressive results on a wide variety of language-related tasks. In this paper, we explore ChatGPT's zero-shot ability to perform affective computing tasks using prompting alone. We show that ChatGPT a) performs meaningful sentiment analysis in the Valence, Arousal and Dominance dimensions, b) has meaningful emotion representations in terms of emotion categories and these affective dimensions, and c) can perform basic appraisal-based emotion elicitation of situations based on a prompt-based computational implementation of the OCC appraisal model. These findings are highly relevant: First, they show that the ability to solve complex affect processing tasks emerges from language-based token prediction trained on extensive data sets. Second, they show the potential of large language models for simulating, processing and analyzing human emotions, which has important implications for various applications such as sentiment analysis, socially interactive agents, and social robotics.
\end{abstract}


\section{Introduction}
Affective Computing focuses on measuring, understanding and impacting human emotion \cite{picard1997,picard2003,calvo2014}. While many approaches have been somewhat successful in doing so, a long-standing challenge in the field has been to appropriately use context for emotional understanding \cite{dudzik2019}. 

Large language models (LLMs), in particular generative pre-trained transformers (GPTs), show impressive results on a wide variety of language-related tasks~\cite{radford2018improving}. This ability is driven by the fact that these models are able to take into account large contexts in the prediction of the next token~\cite{vaswani2017attention}. %, and, are based on a mechanism that allows them to infer word meaning from surrounding words (BERT \cite{devlin2019bert}

Recently, current state-of-the-art GPT-based models allow easy interaction through prompting through the use of user interfaces, effectively providing a human-friendly way of performing zero-shot tasks \cite{ouyang2022training}. And, large scale versions of such models also show impressive results on a diversity of tasks for which they were not fine-tuned~\cite{brown2020language}.

Inspired by these important characteristics of GPT models we investigate the ability to perform affective computing tasks by prompting ChatGPT without any fine-tuning or examples. We address the following research questions and tasks:

\begin{itemize}
\item RQ1: How well can ChatGPT label sentiment, as values on the dimensions Valence, Arousal and Dominance, for textual descriptions of situations 
and for emotion words?
%as well as emotion words?
\item RQ2: To what extent does ChatGPT represent the correct meaning of emotion words and affective dimensions, tested as its ability to map values to situations, and situations to emotion words? 
\item RQ3: How well can ChatGPT predict the emotion class in a situation based on a particular appraisal framework, in our case the Ortony-Clore-Collins (OCC) model \cite{ortony1988cognitive}?
\end{itemize}

We perform a series of exploratory experiments involving prompting ChatGPT to solve the above tasks, as well as comparing the generated results to ground truth data gathered from the literature or through human expert ratings.

\section{Background and related work}

\subsection{Generative Pretrained Transformers}
Generative Pretrained Transformers (GPT) are models trained to generate natural language text, based on the transformer architecture. While the original transformer was designed as an \textit{encoder-decoder} architecture~\cite{vaswani2017attention}, GPT is a decoder-only model: given a prompt it generates text, word by word, based on all the previous context words. The power of transformer models to generate fluent and coherent natural language text stems from the multiple transformer layers with self-attention to compute the strength of the relation between each pair of words in a sequence. This enables the models to capture long-distance dependencies in text~\cite{radford2018improving}. 

The `pretrained' aspect of GPT models refers to the fact that they are pre-trained on huge amounts of text data; for that reason, they are commonly referred to as Large Language Models (LLMs). This pre-training process is based on a language modeling task:
training the model to predict the next word or masked words in a sentence. The first generation of transformer models, including the widely popular encoder models from the BERT-family~\cite{devlin2019bert}, was very successful in a large range of Natural Language Processing (NLP) tasks thanks to the possibility of \textit{fine-tuning} pre-trained models on labeled data.

Since GPT-3, models of the GPT family are capable of solving tasks without fine-tuning: by generating text given a prompt~\cite{brown2020language}. When the prompt does not contain any examples of the correct output (e.g., ``for the sentences below, label the content as having a positive, negative, or neutral sentiment''), we call this zero-shot learning, prompt-based learning, or instruction tuning~\cite{wei2021finetuned}. When the prompt contains a few examples (roughly 3-50) we call this few-shot learning.

In this paper, we experiment with OpenAI's model ChatGPT. It is based on the large pre-trained GPT-3.5 model, finetuned on conversation data, and refined with reinforcement learning based on an expert-trained reward function \cite{ouyang2022training}.

\subsection{Core Affective Computing Tasks}
Affective Computing is a subfield of Artificial Intelligence whose main research includes the detection and interpretation of human affect, on the one hand, \cite{zeng2007,rouast2021}, and the simulation and representation of affect on the other hand in both the human users and the social interactive agents \cite{broekens2021emotion}, see also \cite{calvo2014} and \cite{lugrin2021} for overviews. In this paper, we focus on the use of LLMs for Affective Computing by testing their zero-shot capabilities in automatic affect detection, emotion representation, and the computational elicitation of emotions.

\subsubsection{Automatic affect detection}
Automatic detection and interpretation of behavioral signals of affect involves a broad range of tasks including affect recognition from the face, body, speech, EEG and other physiological signals, and text \cite{zeng2007, poria2017review, rouast2021, chan2023}. As of writing this paper, text is the only modality available in the interaction with ChatGPT. We focus on multidimensional (Valence, Arousal, Dominance) sentiment analysis from short situational descriptions and emotion words.

 Valence refers to the positive-negativeness of an experience, Arousal to the calmness-excitation of that experience, and Dominance to the extent to which a person feels in control of the situation \cite{broekens2012defense,bradley2007affective,russell1977evidence}. While sometimes referred to as Pleasure, we assume here that Valence and Pleasure are similar dimensions both referring to the positiveness-negativeness of an affective state/situation. We use the term Valence, being the "technical" term and avoiding LLM confusion with lexical or common meanings of the word pleasure.

\subsubsection{Affect representation}
Affect representation is typically based on psychological theories of emotion and affect \cite{gebhard2005alma, broekens2021emotion}. 
%In particular, categorical and dimensional representations are used to represent the emotion of an agent, or a human.
In particular, categorical or dimensional representations are used to represent an agent's or human's emotion.
%the emotion of an agent, or the emotion of a human.
%In particular, categorical and dimensional representations are used to represent the emotion of an agent, or the emotion of a human.
We focus on the numerical and latent representation of affect in terms of Valence, Arousal, and Dominance and a mapping from this representation to situations and (categorical) emotion words.

\subsubsection{Computational models of emotion elicitation}
Computational modeling of emotion elicitation is used to simulate affective states based on the appraisal of situations and mental states of an agent. It has been approached from different perspectives, including cognitive agent-based appraisal modeling, e.g. \cite{marsella2009ema}, embodied modeling (homeostasis-based) \cite{canamero2019}, hard-wired appraisal (event coding), and reinforcement learning \cite{broekens2021emotion, broekens2018temporal}. 
In emotion elicitation, the OCC appraisal model \cite{ortony1988cognitive} has arguably had a major influence \cite{tao2005affective}. While the objective of the OCC model was to reverse-engineer a set of emotional experiences based on their cognitive antecedents rather than an exhaustive set of universal rules that represent human appraisal processes (e.g., p.~172), the logical structure of the model has been the basis for a number of computational implementations of the model in Affective Computing \cite{popescu2013gamygdala, gebhard2005alma, dias2005feeling}. As such, in this paper, we focus on the OCC model of appraisal.

\subsection{Related work on sentiment analysis}\label{sec:relwork}
While many sentiment analysis methods are based on supervised fine-tuning (see \cite{chan2023} for a review), there is evidence that generative LLMs are capable of zero-shot prediction tasks, including sentiment classification~\cite{zhong2021adapting,seoh2021open} and argument quality prediction~\cite{van2022will}. It was shown that carefully engineering prompts helps with getting better task-specific guidance~\cite{le2021many}. As such there is reason to believe LLMs will perform on more detailed affect extraction from text as well.

Most work in sentiment focuses on coarse-grained analysis of only Valence (e.g., 2- or 5-point scales) of opinions or reviews, or classification into several basic emotions~\cite{rosenthal2019}. We are interested in a more fine-grained analysis of situations, as this is a more natural test for the understanding of an emotion as experienced by a person. Our tasks/stimuli are therefore different from the sentiment classification tasks such as those used for the SemEval challenges~\cite{rosenthal2019}. We use psychologically validated and fine-grained dimensional ratings of situational descriptions on Valence, Arousal and Dominance \cite{bradley2007affective}.

Recent work investigated the performance of ChatGPT compared to pre-trained and fine-tuned models on a sentiment data set containing tweets with positive and negative labels \cite{amin2023affective}. The work reports reasonable performance as compared to fine-tuned models but only uses a coarse-grained dataset. There is only one prior work that evaluates automatic prediction of sentiment scores using the ANET data. The authors~\cite{mendes2023quantifying} finetune a multi-lingual XLM-RoBERTa-large model and evaluate it on a range of datasets in multiple languages. For the ANET texts (120 items), they report $\rho=0.920$ for Valence and $\rho=0.859$ for Arousal. Dominance was ignored.

\section{Method}
We are interested in the extent to which ChatGPT can perform the three above mentioned affective computing tasks in a zero-shot setting with instruction prompting.

\subsection{Material}
Our data set for RQ 1 consists of all 120 Affective Norms for English Text dataset (ANET) situations \cite{bradley2007affective} and all 151 emotion words from Russel and Mehrabian \cite{russell1977evidence}.

As for RQ2, we focus on mapping stimulus sets to each other, 
with a human expert rater as ground truth.
%and a human expert rater had to also do this for ground truth. 
We used a subset of the words and situations (see Table \ref{anet}, \ref{words}), each containing the 20 most reliably rated items, as assessed by the sum over the squared standard deviation (SD) for Valence, Arousal and Dominance 
(lower values are considered more reliable). 
The SD was given by \cite{bradley2007affective} and \cite{russell1977evidence}.

%For emotion elicitation based on the OCC model, 
For RQ3, we created a set of by-the-authors-formulated situations that reflect each of the tested emotions in the OCC model, see Table \ref{elicitation1}. 
Throughout the experiment, we used the default model of ChatGPT 3.5 between February/March 2023.
%Throughout the entire experiment, we used the default model of ChatGPT 3.5 between February and March 2023. 

\begin{table*}[htbp]
\centering
\caption{Selected ANET situations,  ground truth values for VAD, Generated sentiment incl. failed $\hat{D}*$, explanation in text}
\label{anet}
\begin{tabular}{|c|p{10cm}|c|c|c|c|c|c|c|}
\hline
ID & ANET text & V & A & D & $\hat{V}$ & $\hat{A}$ & $\hat{D}*$ & $\hat{D}$\\
\hline 4650 & You are both aroused, breathless.  You fall together on the couch.  Kisses on your neck, face-- warm hands fumbling with clothing, hearts pounding. & 8.34 & 8.10 & 6.2 & 0.81 & 0.93 & 0.57 & 0.55\\
\hline 2880 & Your friend whispers to you in a meeting, and you strain to catch the words. & 4.63 & 4.96 & 4.4 & 0.5 & 0.43 & 0.27 & 0.39\\
\hline 6820 & You're alone in the alley in a bad part of the city.  A street gang slowly surrounds you, knives out, laughing with menace.  Your heart pounds as they close in. & 1.62 & 8.23 & 1.78 & 0.06 & 0.86 & 0.73 & 0.11\\
\hline 6020 & Without thinking, you stepped off the curb into traffic. Breaks screech. You look up, frozen, heart jumping in your chest.  A truck is skidding, hurtling towards you. & 1.89 & 8.21 & 2.08 & 0.12 & 0.93 & 0.62 & 0.10\\
\hline 5900 & You cringe as a fierce hurricane tears the roof off your house. & 1.81 & 7.98 & 1.83 & 0.05 & 0.63 & 0.64 & 0.23\\
\hline 3310 & You flinch, at the screech of brakes; you look up, and see the speeding car slam into your friend. Her leg is crushed, the artery torn, and blood pumps on the road. & 1.30 & 8.15 & 2.36 & 0 & 0.89 & 0.66 & 0.06\\
\hline 9100 & Clutching his chest, your father falls to the floor, unable to breathe. & 1.30 & 8.13 & 2.23 & 0.12 & 0.71 & 0.48 & 0.05\\
\hline 2540 & You walk through the supermarket aisles checking things off your list as you pick each item you need off the shelves. & 5.54 & 3.38 & 6.7 & 0.63 & 0.29 & 0.24 & 0.5\\
\hline 8040 & Everyone's staring at you, waiting for your presentation.  You've misplaced all your notes, graphics, everything's lost!  What will you say?  They see you shaking, sweating-- mumbling stupidly. & 1.90 & 7.66 & 2.12 & 0.06 & 0.86 & 0.59 & 0.03\\
\hline 8380 & It is a close game, and the crowd cheers as you drive in the winning run. & 8.37 & 8.15 & 7.56 & 0.71 & 0.57 & 0.59 & 0.65\\
\hline 2530 & You've been sick all week, lying on a lumpy couch with a bad cold. & 2.15 & 3.32 & 3.09 & 0.13 & 0.24 & 0.23 & 0.05\\
\hline 7380 & You gag, seeing a roach moving slowly over the surface of the pizza.  You knock the pie on the floor.  Warm cheese spatters on your shoes. & 1.92 & 6.89 & 3.95 & 0.03 & 0.75 & 0.64 & 0.06\\
\hline 4400 & You shiver as your bodies brush together.  You reach out.  You want to touch everywhere, kiss everywhere. You hear the words, "I love you". & 8.28 & 7.91 & 5.90 & 0.87 & 0.75 & 0.46 & 0.64\\
\hline 7040 & You hold the flashlight steady in order to get a better look at the map. & 5.04 & 4.17 & 5.65 & 0.50 & 0.24 & 0.33 & 0.51\\
\hline 8610 & At the net, you go up and block the volleyball perfectly, saving the game. & 8.40 & 7.86 & 7.82 & 0.69 & 0.57 & 0.54 & 0.59\\
\hline 2610 & You are sitting at the kitchen table with yesterday's newspaper in front of you.  You push back the chair when you hear the coffee maker slow to a stop. & 5.37 & 3.13 & 6.11 & 0.40 & 0.14 & 0.21 & 0.47\\
\hline 2130 & You are lying in bed on a Sunday morning, half asleep and listening to the distant sound of bells, relaxing on your day off. & 7.41 & 2.21 & 6.76 & 0.70 & 0.18 & 0.27 & 0.49\\
\hline 8620 & You sprint back to the other side of the court.  You lunge for the ball and hit a spectacular backhand down the line for the win.  You pump your fist in victory. & 8.15 & 7.31 & 8.10 & 0.75 & 0.79 & 0.64 & 0.65\\
\hline 2640 & You laugh so hard with your friends that tears pour down your face. & 8.56 & 7.65 & 5.8 & 0.88 & 0.71 & 0.44 & 0.51\\
\hline 2510 & People are all around you, pressing closer.  It's hard to breath.  You're flushed, sweaty, dizzy, confused.  You realize it's another attack and this time, you think, "I will die." & 1.78 & 7.69 & 2.08 & 0.02 & 0.89 & 0.80 & 0.05\\
\hline & \textbf{Corrs: $N=20$, $p<0.001$ (except failed $\hat{D}$*)} & & &  & \textbf{0.98} & \textbf{0.91} & \textbf{-0.39} & \textbf{0.93}\\
\hline
\end{tabular}
\end{table*}
 
\begin{table}[htbp]
\centering
\caption{Emotion words from Russel with ground truth and extracted sentiment for Pleasure, Arousal, and Dominance}
\label{words}
\begin{tabular}{|p{2cm}|c|c|c|c|c|c|}
\hline  & V & A & D & $\hat{V}$ & $\hat{A}$ & $\hat{D}$ \\
\hline bored & -0.65 & -0.62 & -0.33 & 0.11 & 0.18 & 0.08 \\
\hline triumphant & 0.69 & 0.57 & 0.63 & 0.88 & 0.77 & 0.81 \\
\hline vigorous & 0.58 & 0.61 & 0.49 & 0.76 & 0.77 & 0.8 \\
\hline serious & 0.27 & 0.24 & 0.42 & 0.31 & 0.29 & 0.48 \\
\hline alert & 0.49 & 0.57 & 0.45 & 0.58 & 0.43 & 0.53 \\
\hline astonished & 0.16 & 0.88 & -0.15 & 0.78 & 0.88 & 0.43 \\
\hline uninterested & -0.47 & -0.5 & -0.08 & 0.1 & 0.17 & 0.09 \\
\hline activated & 0.42 & 0.58 & 0.38 & 0.61 & 0.7 & 0.68 \\
\hline enjoyment & 0.77 & 0.44 & 0.42 & 0.91 & 0.79 & 0.67 \\
\hline controlling & 0.47 & 0.34 & 0.66 & 0.16 & 0.43 & 0.84 \\
\hline loved & 0.87 & 0.54 & -0.18 & 0.88 & 0.66 & 0.82 \\
\hline excited & 0.62 & 0.75 & 0.38 & 0.84 & 0.91 & 0.67 \\
\hline influential & 0.68 & 0.4 & 0.75 & 0.46 & 0.49 & 0.81 \\
\hline masterful & 0.58 & 0.44 & 0.69 & 0.62 & 0.61 & 0.82 \\
\hline suspicious & -0.25 & 0.42 & 0.11 & 0.26 & 0.7 & 0.49 \\
\hline mildly annoyed & -0.28 & 0.17 & 0.04 & 0.28 & 0.44 & 0.42 \\
\hline confused & -0.53 & 0.27 & -0.32 & 0.24 & 0.43 & 0.3 \\
\hline friendly & 0.69 & 0.35 & 0.3 & 0.79 & 0.52 & 0.75 \\
\hline aggressive & 0.41 & 0.63 & 0.62 & 0.24 & 0.77 & 0.63 \\
\hline lucky & 0.71 & 0.48 & 0.37 & 0.87 & 0.6 & 0.68 \\
\hline \textbf{Corrs: $N=20$, $p<0.001$} &  &  &  & \textbf{0.77} & \textbf{0.85} & \textbf{0.74} \\
\hline
\end{tabular}
\end{table}

\subsection{Experimental Set-up}
To address our RQs, we performed a series of conversational experiments with the model. To avoid confounding, a new chat session was initiated for each experiment. We explain the process for each of the RQs here.  %, although part of the method.

\subsubsection{RQ1: Sentiment analysis}
For RQ1.1, we prompted\footnote{For readability, prompts are presented alongside the results.} ChatGPT to perform sentiment analysis on Valence, Arousal, and Dominance (VAD), after which the 120 situations were entered as the next prompt (20 per session). The resulting values were correlated (per dimension) with the ground truth values provided by \cite{bradley2007affective}. This serves to test the model's performance on sentiment analysis from situational text.

In RQ1.2, we tested the model's performance on sentiment analysis from emotion words in the same way. These values are correlated with the ground truth provided by \cite{russell1977evidence}. 

\subsubsection{RQ2: Affect representation}
To test the model's affect representation capability, we investigate if the model can use an affective representation in a constructive way.  

For RQ2.1, we repeated RQ1, prompting ChatGPT to assign VAD-values to the 20 most reliable situations and emotion words but now in a single session. Then, we prompted the model to select for each situation the most fitting emotion word, \emph{based on this numerical representation}. We computed the distance matrix between the emotion words and situation stimuli (based on the values generated by ChatGPT) and ranked the selected word according to the distance matrix. This serves to verify if ChatGPT is able to use a \emph{numerical affective representation} to map two stimulus sets to each other. 

For RQ2.2, we repeated the previous setup but provided ChatGPT with the situations and word lists without asking for a numerical representation. We then asked the model to pick for each situation stimulus the two most fitting emotion words. This classification was compared to a ground truth rated by an independent expert in emotion research. This serves to verify if ChatGPT is able to map two stimulus sets to each other, based on a \emph{latent affect representation}. 

For RQ2.3, we prompted ChatGPT to generate a new situation for 9 different value triplets that span the VAD space (1 neutral, 8 in each extreme). The generated situations are classified again by the same independent rater. This serves to verify if ChatGPT is able to generate new situations based on its latent representation of affect, prompted as values on dimensions.

\subsubsection{RQ3: Appraisal-based emotion elicitation}
To assess if ChatGPT can predict emotions according to a specific appraisal framework, we formalized a rule-based logical model of appraisal as a prompt, based on \cite{ortony1988cognitive}. 
The OCC model provides an appraisal structure, including goals as well as global and local variables such as events, agents, and objects in order to describe the elicitation process of a select set of emotions in a rule-based description. 
The OCC authors note that the antecedents of components such as praiseworthiness of actions (i.e standards) and appealingness of objects (i.e. attitudes)
should not be seen as ``internally consistent'' (p.46) and cannot or should not be organized in a representational structure. Thus, for this experiment, we decided to focus only on the branch of event-related emotions of the OCC model.

After formalizing the event-branch of the model in a prompt, we asked ChatGPT to appraise the set of by-the-authors-formulated situations and checked if the corresponding emotion was reported. This serves to assess if ChatGPT can follow a precise logical structure for emotion elicitation, rather than an implicit latent representation: is it `programmable'? 

\section{Results}
\subsection{RQ1: Sentiment analysis}
Pilot testing with prompts on the most reliable subset of situations (the 20 items subset) showed that without an explanation of the Dominance dimension, ChatGPT produces meaningless Dominance values while Valence and Arousal correlate well with the ground truth ($\rho=0.98$ for Valence, $\rho=0.91$ for Arousal and $\rho=-0.39$ for Dominance, see $\hat{D}*$ in Table \ref{anet}). As the correlation is negative, this could indicate confusion about the perspective of who experiences what. We added ``remember that dominance assesses the extent to which the main person in the situation experiences the amount of control it can assert over the situation'' (as taken from \cite{broekens2012defense}). 

As such, for RQ1, we used prompt 1: \begin{quote} 
Valence, Arousal and Dominance are three affective dimensions that you can use to identify the sentiment in sentences. Assume that these dimensions can take values between 0 and 1, with 0 being low, and 1 being high. Remember that dominance assesses the extent to which the main person in the situation experiences the amount of control it can assert over the situation. Assess according to these dimensions the sentiment in the sentences I will give you after. Be precise, and output the values (up until two digits after the decimal point) in a table please. Just acknowledge you got it. [BLOCK OF ANET]\end{quote}
    
The results of the sentiment analysis on all 120 ANET situations (RQ2.1) showed that the Pearson correlations for Valence, Arousal and Dominance with the ground truth data are very strong ($\rho=0.95$, $RMSE=0.08$;  $\rho=0.82$, $RMSE=0.10$; $\rho=0.82$, $RMSE=0.11$, respectively, all $p<0.001$ and $N=120$), indicating a good fit of ChatGPT sentiment to the VAD ground truth values. Correlations for the 20 most reliable situation items are shown in Table \ref{anet}, which are even stronger confirming the higher reliability of the items.

We repeated this process for the emotion words. RQ1.2 results show strong correlations with the ground truth ($\rho=0.89$, $RMSE=0.12$ for Valence, $\rho=0.66$, $RMSE=0.13$ for Arousal, and $\rho=0.68$, $RMSE=0.13$ for Dominance, all $p<0.001$ and $N=151$). Correlations for the 20 most reliable word items are shown in Table \ref{words}.

\subsection{RQ2: Affect representation}
 For RQ2.1, we first prompted ChatGPT with prompt 1 and the 20 item situation list, then with prompt 1 and the list of emotion words. Then we issued prompt 2: \begin{quote}
     Great! now can you use the numerical values of valence, arousal, and dominance to match each sentence from the first list to a word from the second list based on their closeness of values for these affective dimensions?
 \end{quote}
 
 We ranked the selected word for each situation according to a euclidean distance matrix in the VAD space (based on the values given by ChatGPT in this session). The results can be found in Table \ref{Representation1}. The majority of the selected words are amongst the closest distance words, except for situations without a suitable word in the 20 item word list (all fear-related words). This indicates that although ChatGPT is able to map situations to words, it does not seem to use a numerical representation for it, as evidenced by the bad performance on sentences where no suitable word is found semantically, but a much better choice exists numerically.

To test the free-form situation-word mapping (RQ2.2), we repeated the setup of the first part, with a small adjustment to limit ChatGPT hallucinating emotion words that were not part of the list. We prompted each of the ANET situations in an individual session together with the complete list of emotion words to pick from, according to prompt 3: \begin{quote}
[ANET SITUATION] Please pick the two words from this list that fit the situation best based on the affective meaning: [LIST OF EMOTION WORDS]
\end{quote}

\begin{table*}[htbp]
\centering
\caption{Situation Word Mapping with selected word (distance), the ranking of that word according to the distance matrix, the free-form word matching, and the human expert's word matching}
\label{Representation1}
\begin{tabular}{|c|c|c|c|c|}
\hline & \multicolumn{2}{c|}{RQ2.1 Numerical } & \multicolumn{2}{c|}{RQ2.2 latent} \\
\hline ANET & ChatGPT Numerical Mapping (ED) & Rank & ChatGPT's free Mapping & Expert Mapping \\
\hline 4650 & excited(0.08) & 1 & enjoyment, excited & excited, enjoyment  \\
\hline 2880 & confused(0.36) & 2 & alert, mildly annoyed & confused, mildly annoyed  \\
\hline 6820 & aggressive(0.21) & 1 & serious (alert), suspicious & alert, activated\\
\hline 6020 & suspicious(0.45) & 2 & serious (astonished), alert & alert, activated \\
\hline 5900 & vigorous(0.74) & 10 & serious (mildly\_annoyed), astonished & astonished, activated \\
\hline 3310 & astonished(0.67) & 5 & serious (astonished), alert & astonished, alert \\
\hline 9100 & controlling(0.45) & 2 & serious, alert (astonished) & alert, astonished  \\
\hline 2540 & friendly(0.38) & 2 & alert, mildly annoyed (serious) & controlling, serious  \\
\hline 8040 & enjoyment(1.01) & 18 & confused, mildly annoyed (\emph{anxious}) & activated, mildly annoyed  \\
\hline 8380 & triumphant(0.09) & 1 & triumphant, excited & triumphant, vigorous  \\
\hline 2530 & bored(0.23) & 3 & serious (uninterested), mildly annoyed & bored, mildly annoyed  \\
\hline 7380 & excited(0.89) & 12 & astonished (\emph{disgusted}), mildly annoyed & activated, alert  \\
\hline 4400 & loved(0.39) & 7 & excited, loved & loved, excited  \\
\hline 7040 & alert(0.46) & 5 & alert, serious (\emph{focused}) & confused, controlling  \\
\hline 8610 & masterful(0.2) & 3 & triumphant, vigorous (excited) &  masterful, lucky \\
\hline 2610 & bored(0.5) & 5 & alert, mildly annoyed & bored, uninterested  \\
\hline 2130 & lucky(0.5) & 4 & \emph{relaxed}, enjoyment & serious, enjoyment  \\
\hline 8620 & masterful(0.41) & 4 & triumphant, vigorous (excited) & masterful, triumphant  \\
\hline 2640 & enjoyment(0.29) & 6 & enjoyment, friendly & excited, friendly  \\
\hline 2510 & aggressive(0.79) & 5 & confused, serious (\emph{panicked}) & alert, aggressive  \\
\hline
\end{tabular}
\end{table*}

The results for RQ2.2 together with the independent human expert's word selection can be found in Table \ref{Representation1}. For 2 cases a complete match was found, while 11 of the 20 situations had a partial match. Failed matches were mostly due to perspective: the expert rated the feeling of the individual, while ChatGPT rated the situation trying to match both the feeling and the overall affect generally associated with the situation (e.g, most negative situations are rated as `serious'). When we changed the prompt to ``Describe the feeling of the individual experiencing the situation using two words from the following list (you may not use other words):'', we observed 3 complete and 11 partial matches (words in brackets in Table \ref{Representation1}). However, this also pushed the model to consistently generate new better matching words for situations (emphasized words in table) dampening overall matching score. We observe that ChatGPT's capability of mapping two stimulus sets to each other based on latent affect representation seems to match human expert ratings at least in part.

For RQ2.3, we prompted ChatGPT for each of the 9 value triplets in a separate session with the same prompt, varying only VAD-values. We issued prompt 4:
\begin{quote} Assume a man, in a real-live situation that could happen to all of us. Assume that the Valence dimension represents the extent to which the situation is experienced by the man as positive (high) versus negative (low). Assume that the arousal dimension represents the extent to which the situation is experienced by the man as energetic (high) versus calm (low). Assume that the dominance dimension represents the extent to which the man is in control of the situation (high) or not (low). Can you invent a 50 word situation that matches [LOW,HIGH] valence, [LOW,HIGH] arousal, [LOW,HIGH] dominance? \end{quote}

All situations created by ChatGPT were classified as intended by the human expert rater (see Table \ref{Representation3}). This demonstrates that ChatGPT is able to create new emotional situations very reliably when prompted with a specific affective state.

\begin{table*}[htbp]
\centering
\caption{VAD-Octant-based Situation creation}
\label{Representation3}
\begin{tabular}{|c|p{14.5cm}|c|c|}
\hline Prompt & Generated Situation & Rating \\
\hline V+A-D- & The man walks into a peaceful park, surrounded by blooming flowers and chirping birds. He strolls along the path, feeling content and relaxed, taking in the beauty of the scenery, simply existing within it. & V+A-D- \\
\hline V-A+D- & A man is alone at night, walking through a dark and unfamiliar neighborhood. He feels uneasy and senses someone following him, but he cannot see who it is. His heart races and his palms sweat as he quickens his pace. & V-A+D- \\
\hline V-A-D+ & A man walks into a job performance review, expecting a positive evaluation but is met with harsh criticism. He remains calm and composed throughout the meeting, taking control of the conversation and presenting a plan to improve his performance. & V-A-A+ \\
\hline V+A+D- & A man walks into a surprise party where his friends jump out to yell "surprise!" He feels elated and energetic, but also overwhelmed. Although he tries to join in, he feels like a spectator. & V+A+D- \\
\hline V-A+D+ & The man was driving home from work when suddenly a car swerved in front of him, causing him to slam on his brakes and narrowly avoid a collision. Filled with fear and anger, he honks his horn and yells out the window. & V-A+D+ \\
\hline V+A-D+ & A man confidently delivers a well-prepared presentation to a group of colleagues, receiving praise and admiration for his expertise. He maintains a calm and composed demeanor throughout the whole talk. & V+A-D+ \\
\hline V+A+D+ & The man stood at the finish line of a marathon, cheering on his wife as she crossed it, breaking her personal best record. His heart raced with excitement, feeling a surge of pride and accomplishment. As she approached, he raised his arms victoriously. & V+A+D+ \\
\hline V-A-D- & The man trudged through the rain-soaked streets, his soaked shoes squishing with each step. He felt powerless as he searched for shelter, his plans for the day washed away with the downpour. & V-A-D- \\
\hline neutral & A man sits alone in a quiet room, staring out of the window. There are no loud noises or sudden movements to disrupt the peaceful stillness. & neutral \\
\hline
\end{tabular}
\end{table*}

\subsection{RQ3: Appraisal-based emotion elicitation}

\hyphenation{chatOCC}
For the emotion elicitation RQ3, we issued prompt 5: \begin{quote}
    For this conversation, assume we define the chatOCC computational model of emotions as follows. It is based on the OCC (Ortony, Clore, and Collins) appraisal theory. The model defines emotions as a result of cognitive appraisal of an “event” in relation to a specific “goal” a person has in a specific “situation”. We define four appraisals: desirability, certainty of the event, confirmation of the event, and liking of another person. We define the following emotion rules from the perspective of Anne: 
    [RULES FROM TABLE \ref{elicitation1}]

Use ONLY these emotion rules to predict the most likely emotion for the following situation Anne might find herself in, that I will give to you. Be brief, explaining only how the rule matches the situation, and mention the intensity (low, medium, high) based on the desirability and likelihood of the event. Here is the situation: [SITUATION FROM TABLE \ref{elicitation1}] \end{quote}
%Here is the situation: [SITUATION FOR ANNE FROM TABLE \ref{elicitation1}] \end{quote}

For each situation a new session was used. RQ 3 results can be found in Table \ref{elicitation1}. The chatOCC model predicted 10/12 emotions correctly. Two emotions were not correctly identified: Despair (Fears-confirmed) and Disappointment.

\begin{table*}[htbp]
\centering
\caption{Emotion Elicitation}
\label{elicitation1}
\begin{tabular}{|c|p{7cm}|p{7cm}|c|c|}
\hline Emotion & Rule & Situation & Prediction\\
\hline Joy & a desirable event for Anne just happened & Anne just passed her exam. & Joy\\
\hline Distress & an undesirable event for Anne just happened & Anne just failed her exam. & Distress\\
\hline Happy for & a desirable event for a friend of Anne just happened & Bruno just passed his exam. Anne and Bruno get along well. & Happy for\\
\hline Pity & an undesirable event for a friend of Anne just happened & Bruno just failed his exam. Anne and Bruno get along well.
& Pity\\
\hline Gloating & an undesirable event for an enemy of Anne just happened & Bruno just failed her exam. Anne and Bruno do not get along well. & Gloating\\
\hline Resentment & a desirable event for an enemy of Anne just happened. & Bruno just passed his exam. Anne and Bruno do not get along well. & Resentment\\
\hline Hope & a desirable event for Anne might happen in the future. & After studying for weeks, Anne feels ready to take on the exam tomorrow. & Hope\\
\hline Fear & an undesirable event for Anne might happen in the future. & After studying for weeks, Anne feels not ready to take on the exam tomorrow. & Fear\\
\hline Satisfac. & An anticipated desirable event for Anne has indeed happened. & Anne had taken the exam, being sure it would be a success. Checking the results now, she sees that she passed. & Satisfaction\\
\hline Despair & An anticipated undesirable event for Anne has indeed happened. & Anne had taken the exam, being sure it would be a disaster. Checking the results now, she sees that she failed. & Distress\\
\hline Relief & An anticipated undesirable event for Anne did not happen. & Anne had taken the exam, being sure it would be a disaster. Checking the results now, she sees that she passed. & Relief\\
\hline Disapp. & An anticipated desirable event for Anne did not happen. & Anne had taken the exam, being sure it would be a success. Checking the results now, she sees that she failed. & Distress\\
\hline
\end{tabular}
\end{table*}

\section{Discussion}

\subsection{Discussion of RQs}
RQ1.1: The initial negative correlation for Dominance prediction could reflect the issue of perspective-taking discussed in the literature \cite{broekens2012defense}. After correcting this, all correlations between predicted VAD-Values provided by ChatGPT and ground truth values were strong. 
Compared to the correlations reported by~\cite{mendes2023quantifying} (see Section~\ref{sec:relwork}), we see that the predictions by ChatGPT have similar correlations to the fine-tuned XLM-RoBERTa-large model ($\rho=0.92$ for Valence; $\rho=0.86$ for Arousal). This shows that prediction of Valence, Arousal \emph{and Dominance} by generative LLMs is possible without finetuning. 

RQ1.2: The correlations for emotion words are weaker than for situations, but still high and significant. This makes sense and highlights the importance of context for sentiment analysis: situations contain more affective context.

RQ2.1: ChatGPT did not convincingly show the capability to use numerical representations of affect for mapping between stimulus sets but rather seemed to map word meanings to situations. While this proved to work out in some cases (i.e. choosing `triumphant' for situation 8280), in other cases it either failed at representing the affective state from the perspective of the main protagonist (i.e. choosing `excited' for 7380, `aggressive' for 2510), or it got confused because the proper emotion word is not in the list. 
%(i.e. choosing the word `excited' for situation 7380, `aggressive' for situation 2510), or it got confused because the proper emotion word is not in the list. 
This indicates that instead of using the self-created VAD-values for the mapping (as instructed), the model rather seemed to choose emotion words that fit the ANET-situations on an semantic latent level.

RQ2.2: Overall, ChatGPT showed reasonable partial overlap with the human expert's answers. Providing the correct rating perspective increased the overlap, but also increased the tendency to generate new better matching words when no suitable word was in the list. This shows the importance of careful prompting, and, the strong semantic bias such models have. Further, if fear-related words would be added the emotion word list, results are expected to be higher on RQ2.2. We conclude that ChatGPT performs meaningful situation to word mapping based on latent affective representations. 

Results from RQ2.3 show that ChatGPT is indeed able to use 
%numerical 
latent affect representations and validly map these to a newly generated situation. Two major differences in complexity of numerical mapping between RQ2.1 and RQ2.3 are: in RQ2.3 we asked for a first order mapping (VAD-state to stimulus) instead of a second order mapping (situation-to-VAD-state and VAD-state-to-word); and, VAD-states were less specific (low/high vs. exact numerical values). 

RQ3: ChatGPT predicted the majority of emotions correctly given a conceptual framework for appraisal and a situation to appraise. In the two cases that it failed, it incorrectly predicted the less specific emotion of distress instead of despair (aka fears-confirmed) or disappointment. Distress differs from the two latter emotion classes by the relevance of prospect \cite{ortony1988cognitive}. Perhaps ChatGPT did not take the prospect into account when appraising these two cases. However, this seemed only to be the case for negative prospect-based emotions, not for positive ones. It remains unclear why ChatGPT selected distress as the more adequate emotion in this case. However, most of the emotions were predicted correctly indicating a basic capability of applying an appraisal framework for the assessment of given situations and the prediction of emotion classes.

\subsection{Implications for Affective Computing}
\label{TheoreticalImplications}
Our results highlight the interconnection of language and human affect. ChatGPT produces outcomes that, overall, match human data. Apparently, successful use and interpretation of language results in the ability to process affect. Psychological emotion research has suggested language as a crucial, inseparable and highly intertwined context variable for emotion-related processes in human adults \cite{lindquist2013s, barrett2011context}. Although our work cannot be taken as proof for this psychological view of emotion, as LLMs \emph{have} to work from language by default for affect-related tasks, and humans do not necessarily, our work does highlight that language will likely play a crucial role in future Affective Computing research.
%work does highlight that language will likely play a crucial role in Affective Computing research in the future.

This connects to computational models of appraisal theories. While mostly spelled out as logical rules in the respective papers (e.g., \cite{ortony1988cognitive, scherer1984nature, reisenzein2009emotional, smith2009putting}), pioneering work has been put into formalizing these theories into computational frameworks (e.g \cite{popescu2013gamygdala, gebhard2005alma, marsella2009ema, calvo2014}). One critical problem is contextual understanding of the situation to be appraised by the formal model which needs a richly grounded symbolic system. 
%Our work shows the potential of leveraging LLMs for this purpose. 
% SHORTEN: DELETE PARAGRAPH
Symbol grounding is also relevant for making user input interpretable to the computer in the right format. An LLM that interacts with a user in a natural way and extracts formalized interpretations from their input, leaves more room for researchers to focus on problems of affective processing further down the pipeline. Our work shows the potential of leveraging LLMs for this purpose: Symbolic AI is dead, long live symbolic AI!

Finally, for many Affective Computing experiments, sets of stimuli have to be evaluated regarding their affective implications before working with them. This is problematic if the sets are large. While an LLM is not a valid substitute for standardized expert evaluations, given the demonstrated reliability in assigning VAD-values or emotion words to situations, this could help with automatic annotation or creation of tailored or study-customized stimuli -- at least in cases where the affective meaning is not an experiment's main objective. 
%WORKING WITH FINE-TUNED LOCAL LANGUAGE MODELS

\subsection{Limitations and Future Work}
This is an exploratory paper. To generalize these findings, more work is needed in particular using a more diverse collection of datasets for sentiment analysis and a more diverse and more complex set of computational models for appraisal modeling. Further, in this paper we did not address the underlying processing mechanisms of LLMs that enable such powerful zero-shot learning. The results reported in this paper are based on OpenAI's ChatGPT 3.5 from February and March 2023 (which is based on text-davinci-003). New versions are being developed fast. These versions might be more powerful for unseen tasks, but they also might not as the impact of supervised targets and reward-based losses on the generalization to unseen tasks is unknown. Therefore we do not know how generalizable our results are to future LLMs. Another exciting aspect is to explore how LLMs can be used to simulate the dynamics of affect, such as emotion decay and the influence of emotions on mood.

\section{Conclusion}
We show that ChatGPT a) performs accurate extraction of fine-grained, multi-dimensional sentiment from situations and words, compared to the level of fine-tuned models on the same dataset, even on the dimension of Dominance, b) is capable of simple numerical and latent affect representation, and shows moderate understanding of affective dimensions and emotion words, c) can perform basic appraisal-based emotion elicitation of situations based on a prompt-based computational implementation of the OCC appraisal model. These capabilities can be leveraged through appropriate prompting.

These findings are highly relevant: First, we show that the ability to solve complex affect processing tasks emerges from language-based token prediction trained on extensive data sets. Second, we show the potential of large language models for simulating, understanding and analyzing human emotions, which has important implications for various applications such as sentiment analysis, socially interactive agents, and social robotics.
%RUNNING INTO A KNOWLEDGE WALL

\section{Ethical Impact statement}
No subjects were recruited for this work. The environmental impact of the energy used for the inference of the prompts is neglectable due to the small number of prompts. The work could have a major impact on popular belief around psychological capabilities of LLMs, as addressed in the limitations: we explicitly mention this is an exploratory paper. 
%which is addressed in the limitations section: we explicitly mention this is an exploratory paper. 

\section*{Acknowledgments}
This research is partly sponsored by the Hybrid Intelligence project, grant number 024.004.022. Special thanks to Fabiola Diana and her colleagues for their help with data collection.
%[Anonymous to be filled]

\bibliographystyle{IEEEtran}
\bibliography{refs}
\vspace{12pt}
\end{document}